\documentclass[letterpaper, 10 pt, conference]{ieeeconf}  % Comment this line out if you need a4paper

\IEEEoverridecommandlockouts                              % This command is only needed if 
\usepackage{graphicx} % for pdf, bitmapped graphics files
\usepackage{amsmath} % assumes amsmath package installed
\usepackage{amssymb}  % assumes amsmath package installed

\usepackage{multicol}
\usepackage{multirow}
\usepackage{flushend}

\usepackage{ifthen}
\newboolean{preprint} 
\setboolean{preprint}{true}

\makeatletter
\def\ignorecitefornumbering#1{%
     \begingroup
         \@fileswfalse
         #1%                     % do \cite comand
    \endgroup
}
\makeatother

\makeatletter
\let\NAT@parse\undefined
\makeatother
\usepackage[breaklinks,colorlinks,allcolors=blue]{hyperref}

\title{\LARGE \bf
SPOT: Point Cloud Based Stereo Visual Place Recognition for Similar and Opposing Viewpoints
}

\author{Spencer Carmichael$^{1}$, Rahul Agrawal$^{1}$, Ram Vasudevan$^{2}$, and Katherine A. Skinner$^{1}$% <-this % stops a space
\thanks{This work was supported by a grant from Ford Motor Company via the
Ford-UM Alliance under award N028603.}% <-this % stops a space
\thanks{$^{1}$S. Carmichael, R. Agrawal, and K. A. Skinner are with the Department of Robotics, University of Michigan, Ann Arbor, MI 48109 USA
        {\tt\small \{specarmi,rahulagr,kskin\}@umich.edu}}%
        \thanks{$^{2}$R. Vasudevan is with the Department of Robotics and the Department of Mechanical Engineering, University of Michigan, Ann Arbor, MI 48109 USA
       {\tt\small ramv@umich.edu}}%
}

\begin{document}

\maketitle
\thispagestyle{empty}
\pagestyle{empty}

%%%%%%%%%%%%%%%%%%%%%%%%%%%%%%%%%%%%%%%%%%%%%%%%%%%%%%%%%%%%%%%%%%%%%%%%%%%%%%%%
\begin{abstract}

Recognizing places from an opposing viewpoint during a return trip is a common experience for human drivers. However, the analogous robotics capability, visual place recognition (VPR) with limited field of view cameras under 180 degree rotations, has proven to be challenging to achieve. To address this problem, this paper presents Same Place Opposing Trajectory (SPOT), a technique for opposing viewpoint VPR that relies exclusively on structure estimated through stereo visual odometry (VO). The method extends recent advances in lidar descriptors and utilizes a novel double (similar and opposing) distance matrix sequence matching method. We evaluate SPOT on a publicly available dataset with 6.7-7.6 km routes driven in similar and opposing directions under various lighting conditions. The proposed algorithm demonstrates remarkable improvement over the state-of-the-art, achieving up to 91.7\% recall at 100\% precision in opposing viewpoint cases, while requiring less storage than all baselines tested and running faster than all but one. Moreover, the proposed method assumes no \textit{a priori} knowledge of whether the viewpoint is similar or opposing, and also demonstrates competitive performance in similar viewpoint cases. The code is available through the project webpage: \url{https://umautobots.github.io/spot}. 

\end{abstract}

%%%%%%%%%%%%%%%%%%%%%%%%%%%%%%%%%%%%%%%%%%%%%%%%%%%%%%%%%%%%%%%%%%%%%%%%%%%%%%%%
\section{Introduction}

% This ensures the project page is the first citation
\nocite{projectpage}

\begin{figure}
\centering
\includegraphics[width=0.8\columnwidth]{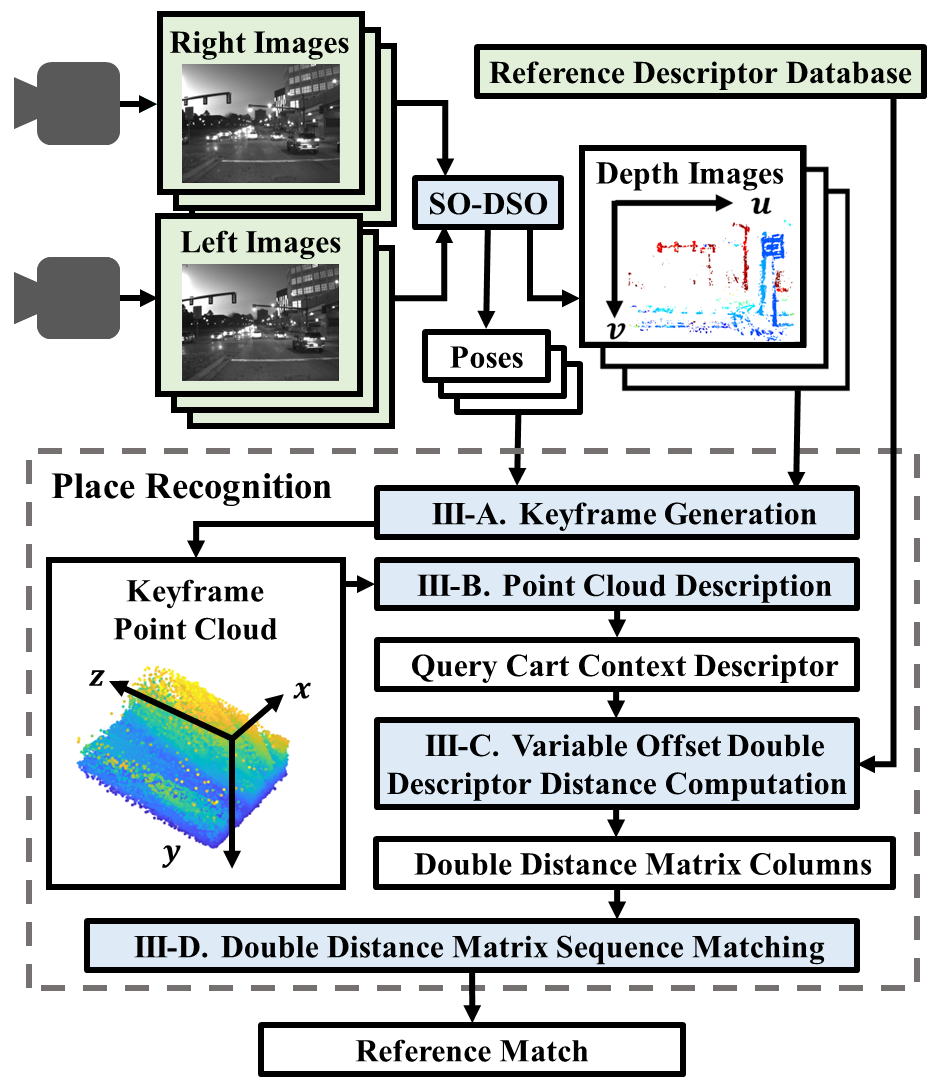}
\caption[Overview of SPOT. External inputs are green, processing blocks are blue, outputs are white. The stereo VO algorithm, SO-DSO, processes stereo images and outputs estimated poses and scaled depth images. This output is accumulated to form a point cloud at a selected keyframe pose. Next, a Cart Context query descriptor is formed. Two distances between the query and all references are computed and a novel double distance matrix sequence matching scheme produces the final reference match.]{Overview of SPOT. External inputs are green, processing blocks are blue, outputs are white. The stereo VO algorithm, SO-DSO \ignorecitefornumbering{\cite{mo_extending_2019}}, processes stereo images and outputs estimated poses and scaled depth images. This output is accumulated to form a point cloud at a selected keyframe pose. Next, a Cart Context \ignorecitefornumbering{\cite{kim_scan_2022}} query descriptor is formed. Two distances between the query and all references are computed and a novel double distance matrix sequence matching scheme produces the final reference match.}  
\label{figure:system}
\vspace{-6mm}
\end{figure}

Visual place recognition (VPR) is an important capability in robotics that supports loop closure, relocalization, and multimap merging \cite{campos_orb-slam3_2021,lowry_visual_2016, zhang_visual_2021}. VPR is challenging because algorithms must avoid false matches between different places with simliar visual appearance while also recognizing revisited places despite possible changes in illumination, viewpoint, weather, seasons, and arrangement of objects (e.g., cars) in the scene. While state-of-the-art methods demonstrate impressive performance in many scenarios \cite{arandjelovic_netvlad_2016, ali2023mixvpr, keetha2023anyloc}, the extreme case of opposing viewpoint VPR is particularly difficult and relatively understudied \cite{keetha2023anyloc, gawel_x-view_2018, garg_dont_2018, garg_lost_2018, garg_look_2019, garg_semanticgeometric_2022}.

Opposing viewpoint VPR involves using cameras with a limited field-of-view (FOV) to perform place recognition under 180 degree rotations. Robustness to 180 degree rotations has been demonstrated in place recognition methods using omnidirectional cameras \cite{arroyo_bidirectional_2014} and lidars \cite{kim_scan_2022}, and 360 degree views fused from multiple cameras \cite{multisphere, xu2023leveraging}. However, such engineering solutions are not without drawbacks. Given the same bandwidth, limited FOV cameras offer higher spatial resolution than omnidirectional cameras, and adding additional cameras increases power consumption, bandwidth, and computing requirements. Opposing viewpoint VPR also offers a fallback in cases where the opposite view is obstructed, such as when a robot is towing a large object or producing a trailing dust cloud. Moreover, existing papers on opposing viewpoint VPR argue that the human capability to solve this problem lends it inherent scientific value \cite{garg_dont_2018, garg_lost_2018, garg_look_2019, garg_semanticgeometric_2022}.

Motivated by the relative invariance of structure to changes in illumination, weather, and seasons, several recent papers have used accumulated point clouds from visual odometry (VO) for VPR rather than \cite{cieslewski_point_2016, ye_place_2017, mo_fast_2020} or in addition to \cite{mo_fast_2020, oertel_augmenting_2020} using appearance. In the case of stereo VO, the estimated point clouds have absolute scale and lidar descriptors can be directly applied \cite{mo_fast_2020}. While these recent papers have demonstrated promising improvement under appearance changes, prior work leveraging VO-derived structure has not explored the opposing viewpoint case.

This paper introduces an opposing viewpoint VPR method that uses limited FOV stereo cameras and relies exclusively on structure estimated through stereo VO (Fig.~\ref{figure:system}). We demonstrate our method's remarkable improvement over the state-of-the-art on the publicly available Novel Sensors for Autonomous Vehicle Perception (NSAVP) dataset \cite{nsavp}. Using a 15 meter localization radius to determine correct matches, our method achieves up to 91.7\% recall at 100\% precision in opposing viewpoint cases while none of the tested baselines exceed 0.2\%. Moreover, our method assumes no \textit{a priori} knowledge of whether the viewpoint is similar or opposing, and also demonstrates competitive performance in similar viewpoint cases. Beyond place recognition performance, we show that our method requires less storage than all baselines tested and runs faster than all but one.

\section{Related Work}

\subsection{Similar Viewpoint VPR}

Place recognition typically proceeds in three stages: a query descriptor is formed, descriptor distances are computed between the query and multiple references, and finally a matching algorithm selects a reference and produces a score. Early VPR techniques used handcrafted local or global image descriptors \cite{lowry_visual_2016}, with methods using local descriptors often employing aggregation schemes such as vector of locally aggregated descriptors (VLAD) \cite{jegou_aggregating_2010}. More recently, VPR has trended towards deep learning-based approaches \cite{zhang_visual_2021, masone_survey_2021}. Inspiration from earlier techniques has led to methods such as NetVLAD \cite{arandjelovic_netvlad_2016}, wherein the output of a CNN is interpreted as dense local descriptors and VLAD is mimicked with a pooling layer. Deep learning-based methods typically demonstrate better performance at the cost of greater computational requirements \cite{zaffar_levelling_2019, zhang_visual_2021}.

Throughout the literature, VPR is often cast as pure image retrieval \cite{lowry_visual_2016}. However, in robotics applications such as SLAM, information relating places is often known and can be exploited \cite{lowry_visual_2016, masone_survey_2021}. For instance, SeqSLAM searches for sequences of query-reference pairs with the knowledge that revisiting an area in the same direction will produce queries in the same order as their matching references \cite{milford_seqslam_2012}. SMART additionally exploits odometry to ensure equi-spaced descriptor formation, improving the search for sequences \cite{pepperell_all-environment_2014, pepperell_towards_2013}.

Recent methods have gone further and leveraged VO-derived structure \cite{cieslewski_point_2016, ye_place_2017, mo_fast_2020, oertel_augmenting_2020}, effectively performing video retrieval \cite{garg_where_2021}. Most similar to our work, \cite{mo_fast_2020} accumulates the output of SO-DSO \cite{mo_extending_2019}, a stereo extension of DSO \cite{engel_direct_2018}, and directly applies lidar descriptors for place recognition. Among the lidar descriptors tested, Scan Context \cite{kim_scan_2018} demonstrated the best performance and computational efficiency \cite{mo_fast_2020}. Although Scan Context is invariant to rotations \cite{kim_scan_2018}, its potential for opposing viewpoint VPR was not investigated \cite{mo_fast_2020}.

Inspired by \cite{mo_fast_2020}, we similarly utilize SO-DSO \cite{mo_extending_2019} to estimate structure and apply an updated version of Scan Context, called Cart Context \cite{kim_scan_2022}, for place description. To further enhance performance, we introduce an improved Cart Context distance metric and a novel double distance matrix sequence matching method, and use pose estimates to ensure equi-spaced descriptors as in SMART \cite{pepperell_all-environment_2014, pepperell_towards_2013}.

\subsection{Opposing Viewpoint VPR}\label{section:opposing_viewpoint_vpr}

Opposing viewpoint VPR methods have mainly involved leveraging semantic information with single-image descriptors \cite{garg_lost_2018, garg_dont_2018, garg_look_2019, garg_semanticgeometric_2022}. In \cite{garg_lost_2018}, the Local Semantic Tensor (LoST) descriptor was introduced, which is built from both the final output of a dense semantic segmentation network and the output of one of its intermediate layers (conv5). The full method, LoST-X, adds a step to filter the top nearest neighbor candidates by again utilizing the semantic predictions and conv5 features \cite{garg_lost_2018}. Improved variations of the LoST-X framework were presented in two subsequent papers \cite{garg_look_2019, garg_semanticgeometric_2022}. In \cite{garg_look_2019}, LoST is replaced by NetVLAD and the fine matching stage is updated to consider candidate reference \textit{sequences} and to leverage learning-based single-view depth estimates \cite{garg_look_2019}. This method, sequence-to-single (Seq2single), uses depth estimates only to filter keypoints \cite{garg_look_2019}, unlike our method where depth estimates form the basis of the place description. Most recently, AnyLoc has demonstrated universal place recognition, encompassing diverse viewpoints, with single-image descriptors formed by aggregating features from a foundation model \cite{keetha2023anyloc}. However, AnyLoc's opposing viewpoint capability was not evaluated outdoors or against methods specific to this task \cite{keetha2023anyloc}. 

In approaches with single-image descriptors, matches are often separated by a \textit{visual offset}: the physical distance between two cameras with opposing viewpoints at which the visual overlap between them is maximized \cite{garg_semanticgeometric_2022}. This distance, estimated to be 30-40 meters in a driving scenario \cite{garg_semanticgeometric_2022}, may be difficult to overcome in applications where VPR must be followed by precise relative pose estimation (e.g., loop closure in SLAM). Better localization accuracy has been shown with semantic graphs built across multiple images, but this was only demonstrated with short ($<$1 km) sequences \cite{gawel_x-view_2018}. Existing methods also suffer from additional drawbacks, such as required \textit{a priori} knowledge of whether the viewpoint is similar or opposing \cite{garg_lost_2018, garg_semanticgeometric_2022}, high storage demands \cite{garg_lost_2018, garg_look_2019, garg_semanticgeometric_2022}, and the resource requirements imposed by deep neural networks \cite{keetha2023anyloc, gawel_x-view_2018, garg_dont_2018, garg_lost_2018, garg_look_2019, garg_semanticgeometric_2022}. Our proposed method, SPOT, avoids the visual offset problem by forming place descriptors that capture the structure \textit{surrounding} the camera rather than being limited to the FOV of a single image. SPOT is also fully handcrafted, runs quickly on a CPU, requires no \textit{a priori} knowledge of whether the viewpoint is similar or opposing, and requires relatively little storage for the reference database.

\begin{figure*}
\centering
\includegraphics[width=0.97\textwidth]{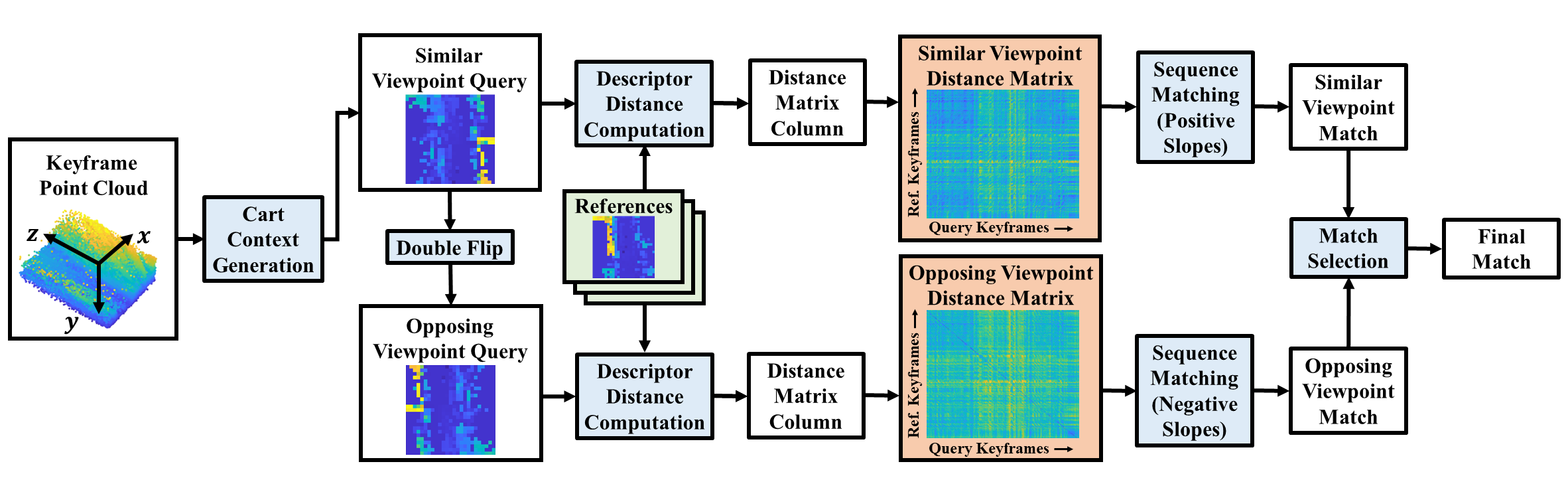}
\caption{An expanded depiction of the final three place recognition stages. External inputs are green, processing blocks are blue, outputs are white, and incrementally updated objects are orange. A Cart Context \cite{kim_scan_2022} descriptor is formed from the most recent keyframe point cloud and flipped about both axes to produce an additional descriptor for opposing viewpoint VPR. Descriptor distances are computed between each query and every reference to produce two separate distance matrices for similar and opposing viewpoints. Sequence matching, as described in \cite{milford_seqslam_2012}, is performed separately in each distance matrix and the final output is selected from the results. \vspace{-4mm}}
\label{figure:place_recognition}
\end{figure*}
\section{Technical Approach}

Figure~\ref{figure:system} shows an overview of the full SPOT system, which takes as input stereo images and outputs place recognition matches at selected keyframes.
The entry point of the system is a stereo VO algorithm, which outputs poses and sparse depth images with absolute scale. Any stereo VO algorithm could be used for this purpose. We choose SO-DSO \cite{mo_extending_2019} due to its demonstrated success at generating point clouds useful for VPR \cite{mo_fast_2020}.
The following subsections describe the remaining stages of the algorithm. Figure~\ref{figure:place_recognition} provides an expanded depiction of the final three stages.

\subsection{Keyframe Generation}\label{section:keyframe_generation}

The keyframe generation component serves to accumulate the output of SO-DSO, select when a keyframe should be created, and generate point clouds for Cart Context \cite{kim_scan_2022} description. Stereo triangulation error increases rapidly with depth \cite{matthies_error_1987}, so as an initial step to reject noisy points, we discard all pixels of the depth image exceeding a given threshold, $r_d$. Each depth image from SO-DSO has a corresponding pose estimate. We use this pose estimate and the known intrinsics of the left camera to project each valid pixel of the depth image into a common world frame. We continuously accumulate projected points from each new depth image into a single point cloud. As in SMART \cite{pepperell_towards_2013, pepperell_all-environment_2014}, we aim to create place descriptors at a constant distance apart to aid subsequent sequence matching. To do this, we compute the path distance, i.e., the sum of Euclidean distances between SO-DSO position estimates, since the last keyframe. When this path distance exceeds the desired descriptor spacing, $s$, we produce a new keyframe point cloud centered at the current pose. To create a new keyframe point cloud, we transform the accumulated point cloud into the current camera frame and select all points residing within a horizontal radius $r_k$ about the camera position. Every time a new keyframe is generated, we cull faraway points in the accumulated point cloud using a second, larger horizontal radius, $r_a$. This culling improves efficiency and accounts for large-scale odometry drift. We do not create the first keyframe point cloud until the path distance exceeds $1.5r_k$ in an effort to ensure the point cloud is well-populated.

\subsection{Point Cloud Description}\label{section:point_cloud_description}

For each keyframe point cloud, a descriptor is formed. As the depth estimates obtained through SO-DSO have absolute scale, it is possible to directly apply a lidar descriptor. We choose to use the Cart Context descriptor introduced in \cite{kim_scan_2022} as it is highly efficient to compute and demonstrates impressive performance in lidar-based place recognition.

The Cart Context descriptor is created from the keyframe points lying in a $2r_{lo}\times 2r_{la}$ meter horizontal rectangle centered at the origin of the camera frame, with the $2r_{lo}$ meter side aligned with the longitudinal direction (or forward direction, $z$) and the $2r_{la}$ meter side aligned with the lateral direction ($x$). The rectangular domain is divided into $m$ equal-sized rows along the longitudinal axis and $n$ equal-sized columns along the lateral axis to create bins. The maximum height above the ground of the points captured within each bin is recorded to obtain the $m\times n$ Cart Context descriptor. If no point exists within a bin, the corresponding value in the descriptor is set to zero. The height above the ground of a point, $\mathbf{p} = [x, y, z]^T$, is computed as $h_p = h_c - y$, where $h_c$ is the known height of the left camera above the ground. Examples of Cart Context descriptors are visualized in Fig.~\ref{figure:place_recognition}. To ensure the rectangular domain of the Cart Context descriptor is fully populated with points, the following relationships should be satisfied: $r_{d} \geq r_{lo}$ and $r_k \geq \sqrt{r_{lo}^2 + r_{la}^2}$. 

\subsection{Variable Offset Double Descriptor Distance Computation}\label{section:variable_offset_double_descriptor_distance_computation}

\begin{figure}
\centering
\includegraphics[width=0.42\columnwidth]{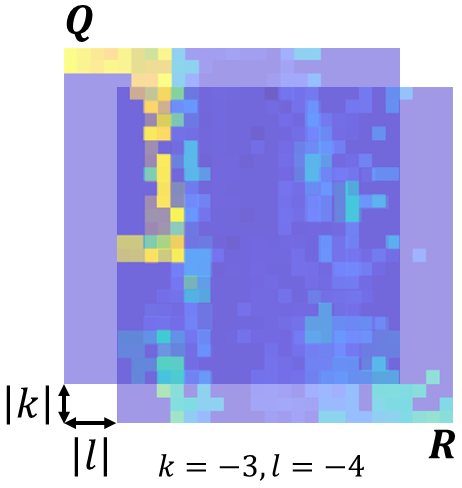}
\caption{A depiction, for a single longitudinal ($k$) and lateral ($l$) shift, of the overlapping regions of the query $\mathbf{Q}$ and reference $\mathbf{R}$ between which the cosine distance is computed. \vspace{-6mm}}  
\label{figure:variable_offset}
\end{figure}

The Cart Context descriptor offers a coarse, birds-eye-view representation of the structure surrounding the keyframe camera pose. In \cite{kim_scan_2022}, the Cart Context descriptor distance is computed as the minimum column-wise cosine distance between the reference and all circular column shifts of the query. The circular shifts lend the distance computation lateral robustness but have no valid physical interpretation. We instead compute the descriptor distance using the variable offset concept previously applied in SMART \cite{pepperell_towards_2013, pepperell_all-environment_2014}. While Sum of Absolute Differences (SAD) is applied to the overlapping patches in SMART, we have empirically observed the best performance using cosine distance on the flattened patches. Specifically, the descriptor distance is computed as:
\begin{equation}\label{equation:descriptor_distance}
\begin{split}
    d(\mathbf{Q}, \mathbf{R}) & = \min_{k\in s_{lo}, l\in s_{la}} \text{cd}\left( \mathbf{Q}[i_Q, j_Q, h, w], \mathbf{R}[i_R, j_R, h, w] \right) \\
    & i_Q = \max(1, -k + 1),\ j_Q = \max(1, -l + 1) \\
    & i_R = \max(1, k + 1),\ j_R = \max(1, l + 1) \\
    & h = m - |k|,\ w = n - |l|
\end{split}
\end{equation}
where $\mathbf{Q}\in\mathbb{R}^{m\times n}$ and $\mathbf{R}\in\mathbb{R}^{m\times n}$ are the query and reference Cart Context descriptors, $\mathbf{Q}[i,j,h,w]$ denotes the submatrix obtained by selecting the rows $\{i, ...,i + h - 1\}$ and columns $\{j, ..., j + w - 1\}$ from $\mathbf{Q}$, $s_{lo}$ and $s_{la}$ are sets of longitudinal and lateral shifts, and $\text{cd}(\mathbf{A}, \mathbf{B})$ is the cosine distance between flattened matrices $\mathbf{A}$ and $\mathbf{B}$. Figure~\ref{figure:variable_offset} visualizes how the longitudinal and lateral shifts are applied to the query and reference descriptors in Equation~\ref{equation:descriptor_distance}.

The descriptor distance procedure described above results in robustness to longitudinal and lateral shifts but not rotations. In \cite{kim_scan_2022}, the opposing viewpoint case is accounted for by double flipping the reference descriptors, i.e., one flip about each axis. The double flipped versions are then treated as additional reference descriptors corresponding to the same places as their non-flipped counterparts \cite{kim_scan_2022}. This method is referred to as Augmented Cart Context (A-CC) \cite{kim_scan_2022}. The method works because the double flip intuitively yields a descriptor similar to that which would have been produced from the opposing view. Here, we instead perform the double flip on the query descriptor rather than the references to avoid either doubling the reference database size or requiring that the references be double flipped for each new query. As depicted in Fig.~\ref{figure:place_recognition}, for each new query and its double flipped counterpart, descriptor distances are computed against every reference in the database. For efficiency, computations across separate references are performed in parallel. 

\subsection{Double Distance Matrix Sequence Matching}\label{section:double_distance_matrix_sequence_matching}

The descriptor distances computed in the previous stage contribute one new column each to two separate, continuously updated distance matrices. The descriptor distances computed with the original query contribute to a distance matrix that captures similar viewpoint matches $\mathbf{D}_{sim}$, while those computed with the double flipped query contribute to a distance matrix that captures opposing viewpoint matches $\mathbf{D}_{opp}$. We expect a sequence of similar viewpoint matches to appear as a line with positive slope in $\textbf{D}_{sim}$ and to produce no pattern in $\mathbf{D}_{opp}$. Conversely, we expect a sequence of opposing viewpoint matches to appear as a line with negative slope in $\mathbf{D}_{opp}$ and to produce no pattern in $\textbf{D}_{sim}$. Note that the equi-spaced descriptor formation described in Section~\ref{section:keyframe_generation} should ensure a slope magnitude roughly equal to 1 in either case. The distance matrices shown in Fig.~\ref{figure:place_recognition} were computed in an opposing viewpoint scenario and illustrate these expectations.

To predict the correct match without \textit{a priori} knowledge of whether the viewpoint is similar or opposing, we perform sequence matching, as described in \cite{milford_seqslam_2012}, separately within each distance matrix. Specifically, over the last $w$ queries we sum over lines in $\mathbf{D}_{sim}$ with positive slopes (referred to as velocities in \cite{milford_seqslam_2012}) and in $\mathbf{D}_{opp}$ with negative slopes, searching for a line with the minimum sum. In each case, we evaluate slopes with magnitudes ranging from $v_{min}$ to $v_{max}$. For efficiency, sums over multiple candidate lines are computed in parallel. Each of the two searches returns a predicted match for the query at the center of the search window. The match with the lowest sum is the final output and its score is computed as the lowest sum divided by the second lowest sum outside of a window centered on the match.

\section{Experimental Setup}

\subsection{Dataset Overview}\label{section:dataset_overview}

\begin{figure}
\centering
\includegraphics[width=1.0\columnwidth]{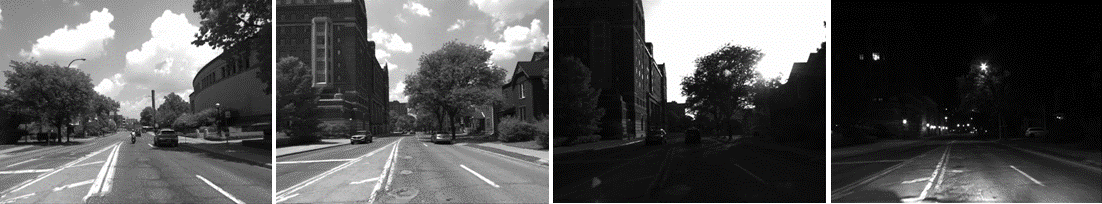}
\caption{Opposing viewpoint images from route \textit{R0} \cite{nsavp} at the same place. \textbf{Left to right:} noon reference, noon query, sunset query and night query.}
\label{figure:lighting_conditions}
\end{figure}

\begin{figure}
\centering
\includegraphics[width=1.0\columnwidth]{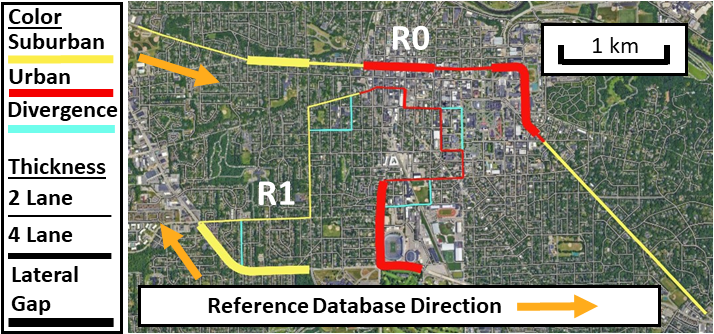}
\caption{Evaluated routes from the NSAVP dataset.}  
\label{figure:route}
\vspace{-4mm}
\end{figure}

The Oxford RobotCar dataset \cite{maddern_1_2017} is commonly used for evaluating opposing viewpoint VPR between front- and rear-facing cameras \cite{garg_dont_2018, garg_lost_2018, garg_look_2019, garg_semanticgeometric_2022}. However, while the Oxford Robotcar dataset includes a front-facing stereo camera, it has only short segments driven in opposing directions \cite{maddern_1_2017}. Therefore, we instead utilize the NSAVP dataset \cite{nsavp}, which includes front-facing stereo monochrome and RGB cameras and $\sim$8 km routes, \textit{R0} and \textit{R1}, driven fully in opposing lanes. The sequences capture a variety of lighting conditions (Fig.~\ref{figure:lighting_conditions}), scene types (urban vs. suburban), traffic conditions, road widths (two vs. four lanes), and lateral gaps between lanes (e.g., medians and turn lanes). We use 6.7 and 7.6 km subsets of the \textit{R0} and \textit{R1} routes respectively (Fig.~\ref{figure:route}). We form reference descriptor databases for each route by applying the steps described in Sections~\ref{section:keyframe_generation} and \ref{section:point_cloud_description} to sequences collected at noon. Query sequences will be referred to by their route, time of day, and viewpoint relative to the reference database. The one exception is a \textit{R1} sequence collected at noon with divergences from its route, which will be referred to as \textit{Diverge R1} (these divergences are shown in Fig.~\ref{figure:route})\footnote{Original sequence names \cite{nsavp}: references: R0\_RA0, R1\_RA0; queries: R0\_FA0, R0\_FS0, R0\_FN0, R1\_FA0, R1\_DA0, R0\_RS0, R0\_RN0.}.

\subsection{Implementation Details}

We run SO-DSO on the monochrome images of all sequences with default parameters. Following \cite{oertel_augmenting_2020}, we discard images during periods where the vehicle is stationary to ensure stable tracking. The parameters used for SPOT are listed in Table~\ref{table:parameters}. Despite the NSAVP dataset only containing lateral offsets in one direction, we use a symmetric set of lateral shifts to avoid \textit{a priori} assumptions regarding lane shifts. We additionally conducted ablation and sequence length studies which are \ifthenelse{\boolean{preprint}}{presented in appendices \ref{appendix:ablation_study} and \ref{appendix:sequence_study}}{included in the appendix \cite{projectpage}}.

\begin{table}[]
\centering
\caption{SPOT parameter values.}
\resizebox{1.0\columnwidth}{!}{%
\begin{tabular}{|l||l|l|}
\hline
\textbf{Parameter(s)} & \textbf{Value(s)} & \textbf{Description}        \\ \hline\hline
$r_d$, $r_k$, $r_a$   & 35.35 m, 35.35 m, 90 m  & Point cloud thresholds                \\ \hline
$s$                & 2 m     & Keyframe spacing              \\ \hline
$r_{lo}$, $r_{la}$          & 25 m, 25 m  & Descriptor ranges               \\ \hline
$m$, $n$                & 25, 25   & Descriptor dimensions                 \\ \hline
$s_{lo}$           & \{-2, 1, 0, 1, 2\}  & Variable offset longitudinal shifts  \\ \hline
$s_{la}$           & \{-5, -4, ..., 4, 5\} & Variable offset lateral shifts \\ \hline
$w$                & 75    & Sequence length                \\ \hline
$v_{min}$, $v_{max}$          & 0.6, 1.4  & Sequence matching slope range                  \\ \hline
\end{tabular}%
}
\label{table:parameters}
\vspace{-2.5mm}
\end{table}

\subsection{Baselines}\label{section:baselines}
We evaluate our method against two state-of-the-art opposing viewpoint VPR frameworks: LoST-X \cite{garg_lost_2018} and Seq2single \cite{garg_look_2019}. Following \cite{garg_lost_2018}, we additionally test LoST, LoST-X, and NetVLAD \cite{arandjelovic_netvlad_2016} with sequence matching using OpenSeqSLAM \cite{sunderhauf_are_nodate} (denoted with +SM). We use the same sequence length as used in SPOT. To align with \cite{garg_lost_2018} and \cite{garg_look_2019}, we use ground truth position data to sample the left RGB images input to these methods at a constant 2 meter distance apart. Note that this sampling lends these baselines an advantage over our proposed approach, as SPOT selects equi-spaced keyframes based on VO pose estimates, rather than ground truth. We also test the method proposed in \cite{mo_fast_2020}, referred to here as SO-DSO VPR. This method also employs a rotation-invariant Scan Context descriptor with SO-DSO derived points clouds, but it was not originally evaluated against opposing viewpoints.

\subsection{Evaluation Methodology \& Metrics}\label{section:evaluation_and_metrics}

\begin{table}[]
\centering
\caption{Labels applied to each query}
\resizebox{0.8\columnwidth}{!}{%
\begin{tabular}{|llll|}
\hline
\multicolumn{4}{|l|}{\textbf{If query $i$ returns an accepted reference match $j$}}                                                                                                              \\ \hline
\multicolumn{1}{|l|}{If $|\mathbf{c}_i - \mathbf{c}_j| \leq r_m$}                                                                                & \multicolumn{3}{l|}{true positive}  \\ \hline
\multicolumn{1}{|l|}{Otherwise}                                                                                                                & \multicolumn{3}{l|}{false positive} \\ \hline\hline
\multicolumn{4}{|l|}{\textbf{If query $i$ returns no accepted match}}                                                                                                                     \\ \hline
\multicolumn{1}{|l|}{If $\exists j' \in J, |\mathbf{c}_i - \mathbf{c}_{j'}| \leq r_m$ } & \multicolumn{3}{l|}{false negative} \\ \hline
\multicolumn{1}{|l|}{Otherwise}                                                                                                                & \multicolumn{3}{l|}{true negative}  \\ \hline
\multicolumn{2}{l}{${c}_{i}$, ${c}_{j}$, and ${c}_{j'}$ are ground truth positions} \\
\multicolumn{2}{l}{$r_m$ is a localization radius} \\
\multicolumn{2}{l}{$J$ is the set of all reference indices}
\end{tabular}%
}
\label{table:definitions}
\vspace{-6mm}
\end{table}

We consider a query $i$ to return an accepted reference match $j$ if the match score is less than or equal to a given threshold. Note that sequence matching produces no match for the first and last $(w - 1)/2$ queries \cite{milford_seqslam_2012}, and we consider such queries to return no accepted match by default. For a given score threshold, precision and recall are computed by assigning a label to each query according to Table~\ref{table:definitions}. Precision-recall (PR) curves are drawn by varying the score threshold from the minimum to the maximum match score across all queries. We compute PR curves using two different values for the localization radius, $r_m$: 15 meters and 80 meters. The larger 80 meter threshold is chosen to match that used in \cite{garg_lost_2018} and the stricter 15 meter threshold is chosen to be just above the maximum lateral shift between opposing lanes in the NSAVP dataset (12.5 meters).

From the PR curves, we compute two metrics: maximum recall at 100\% precision (MR100) and the area under the PR curve (AUC). MR100 indicates the percentage of all possible matches successfully attained without any false positives, while the AUC provides a summary of the performance that is less sensitive to individual false positives. In the ideal case, both metrics would equal 1. For methods utilizing sequence matching, the maximum possible value for both metrics is less than 1 due to the queries that are not assigned a match, so results with zero incorrect matches are denoted.

\section{Results}

\subsection{Place Recognition Performance}\label{section:pr_performance}

\begin{table*}[]
\centering
\caption{Place recognition performance: MR100 and AUC using a 15 meter and 80 meter localization radius}
\resizebox{1.00\textwidth}{!}{
\begin{tabular}{ccccccccccccccc}
\hline
\multicolumn{1}{|c||}{\multirow{3}{*}{\textbf{Method}}}  & \multicolumn{10}{c||}{\textbf{Opposing Viewpoint}}  & \multicolumn{4}{c|}{\textbf{Similar Viewpoint}}
\\ \cline{2-15}
\multicolumn{1}{|l||}{}  & \multicolumn{2}{c|}{\textit{\textbf{Noon R0}}}  & \multicolumn{2}{c|}{\textit{\textbf{Sunset R0}}}  & \multicolumn{2}{c|}{\textit{\textbf{Night R0}}}  & \multicolumn{2}{c|}{\textit{\textbf{Noon R1}}}  & \multicolumn{2}{c||}{\textit{\textbf{Diverge R1}}}  & \multicolumn{2}{c|}{\textit{\textbf{Sunset R0}}}  & \multicolumn{2}{c|}{\textit{\textbf{Night R0}}}
\\ \cline{2-15}
\multicolumn{1}{|l||}{}  & \multicolumn{1}{c|}{\textbf{15 m}}  & \multicolumn{1}{c|}{\textbf{80 m}}  & \multicolumn{1}{c|}{\textbf{15 m}}  & \multicolumn{1}{c|}{\textbf{80 m}}  & \multicolumn{1}{c|}{\textbf{15 m}}  & \multicolumn{1}{c|}{\textbf{80 m}}  & \multicolumn{1}{c|}{\textbf{15 m}}  & \multicolumn{1}{c|}{\textbf{80 m}}  & \multicolumn{1}{c|}{\textbf{15 m}}  & \multicolumn{1}{c||}{\textbf{80 m}}  & \multicolumn{1}{c|}{\textbf{15 m}}  & \multicolumn{1}{c|}{\textbf{80 m}}  & \multicolumn{1}{c|}{\textbf{15 m}}  & \multicolumn{1}{c|}{\textbf{80 m}}
\\ \hline\hline
\multicolumn{15}{|c|}{MR100}
\\ \hline
\multicolumn{1}{|l||}{\textbf{SPOT (Ours)}}  & \multicolumn{1}{c|}{\textbf{0.917}}  & \multicolumn{1}{c|}{\textbf{0.913}}  & \multicolumn{1}{c|}{\textbf{0.836}}  & \multicolumn{1}{c|}{\textbf{0.832}}  & \multicolumn{1}{c|}{\textbf{0.067}}  & \multicolumn{1}{c|}{\textbf{0.067}}  & \multicolumn{1}{c|}{\textbf{0.798}}  & \multicolumn{1}{c|}{\textbf{0.809}}  & \multicolumn{1}{c|}{\textbf{0.605}}  & \multicolumn{1}{c||}{\textbf{0.536}}  & \multicolumn{1}{c|}{0.968*}  & \multicolumn{1}{c|}{0.968*}  & \multicolumn{1}{c|}{0.199}  & \multicolumn{1}{c|}{0.199}
\\ \hline
\multicolumn{1}{|l||}{\textbf{LoST-X}}  & \multicolumn{1}{c|}{0.000}  & \multicolumn{1}{c|}{0.001}  & \multicolumn{1}{c|}{0.000}  & \multicolumn{1}{c|}{0.000}  & \multicolumn{1}{c|}{0.000}  & \multicolumn{1}{c|}{0.000}  & \multicolumn{1}{c|}{0.000}  & \multicolumn{1}{c|}{0.004}  & \multicolumn{1}{c|}{0.000}  & \multicolumn{1}{c||}{0.007}  & \multicolumn{1}{c|}{0.819}  & \multicolumn{1}{c|}{0.900}  & \multicolumn{1}{c|}{0.018}  & \multicolumn{1}{c|}{0.018}
\\ \hline
\multicolumn{1}{|l||}{\textbf{LoST+SM}}  & \multicolumn{1}{c|}{0.000}  & \multicolumn{1}{c|}{0.010}  & \multicolumn{1}{c|}{0.000}  & \multicolumn{1}{c|}{0.009}  & \multicolumn{1}{c|}{0.000}  & \multicolumn{1}{c|}{0.006}  & \multicolumn{1}{c|}{0.000}  & \multicolumn{1}{c|}{0.056}  & \multicolumn{1}{c|}{0.000}  & \multicolumn{1}{c||}{0.049}  & \multicolumn{1}{c|}{\textbf{0.978}*}  & \multicolumn{1}{c|}{\textbf{0.978}*}  & \multicolumn{1}{c|}{0.309}  & \multicolumn{1}{c|}{\textbf{0.335}}
\\ \hline
\multicolumn{1}{|l||}{\textbf{LoST-X+SM}}  & \multicolumn{1}{c|}{0.000}  & \multicolumn{1}{c|}{0.036}  & \multicolumn{1}{c|}{0.000}  & \multicolumn{1}{c|}{0.015}  & \multicolumn{1}{c|}{0.000}  & \multicolumn{1}{c|}{0.021}  & \multicolumn{1}{c|}{0.000}  & \multicolumn{1}{c|}{0.100}  & \multicolumn{1}{c|}{0.000}  & \multicolumn{1}{c||}{0.041}  & \multicolumn{1}{c|}{\textbf{0.978}*}  & \multicolumn{1}{c|}{\textbf{0.978}*}  & \multicolumn{1}{c|}{\textbf{0.330}}  & \multicolumn{1}{c|}{0.330}
\\ \hline
\multicolumn{1}{|l||}{\textbf{Seq2single}}  & \multicolumn{1}{c|}{0.000}  & \multicolumn{1}{c|}{0.000}  & \multicolumn{1}{c|}{0.000}  & \multicolumn{1}{c|}{0.000}  & \multicolumn{1}{c|}{0.000}  & \multicolumn{1}{c|}{0.000}  & \multicolumn{1}{c|}{0.000}  & \multicolumn{1}{c|}{0.002}  & \multicolumn{1}{c|}{0.000}  & \multicolumn{1}{c||}{0.002}  & \multicolumn{1}{c|}{0.166}  & \multicolumn{1}{c|}{0.195}  & \multicolumn{1}{c|}{0.000}  & \multicolumn{1}{c|}{0.008}
\\ \hline
\multicolumn{1}{|l||}{\textbf{NetVLAD+SM}}  & \multicolumn{1}{c|}{0.000}  & \multicolumn{1}{c|}{0.029}  & \multicolumn{1}{c|}{0.000}  & \multicolumn{1}{c|}{0.019}  & \multicolumn{1}{c|}{0.000}  & \multicolumn{1}{c|}{0.003}  & \multicolumn{1}{c|}{0.000}  & \multicolumn{1}{c|}{0.000}  & \multicolumn{1}{c|}{0.000}  & \multicolumn{1}{c||}{0.000}  & \multicolumn{1}{c|}{\textbf{0.978}*}  & \multicolumn{1}{c|}{\textbf{0.978}*}  & \multicolumn{1}{c|}{0.269}  & \multicolumn{1}{c|}{0.269}
\\ \hline
\multicolumn{1}{|l||}{\textbf{SO-DSO VPR}}  & \multicolumn{1}{c|}{0.000}  & \multicolumn{1}{c|}{0.000}  & \multicolumn{1}{c|}{0.000}  & \multicolumn{1}{c|}{0.008}  & \multicolumn{1}{c|}{0.000}  & \multicolumn{1}{c|}{0.000}  & \multicolumn{1}{c|}{0.002}  & \multicolumn{1}{c|}{0.002}  & \multicolumn{1}{c|}{0.000}  & \multicolumn{1}{c||}{0.000}  & \multicolumn{1}{c|}{0.111}  & \multicolumn{1}{c|}{0.111}  & \multicolumn{1}{c|}{0.000}  & \multicolumn{1}{c|}{0.000}
\\ \hline\hline
\multicolumn{15}{|c|}{AUC}
\\ \hline
\multicolumn{1}{|l||}{\textbf{SPOT (Ours)}}  & \multicolumn{1}{c|}{\textbf{0.970}}  & \multicolumn{1}{c|}{\textbf{0.965}}  & \multicolumn{1}{c|}{\textbf{0.968}}  & \multicolumn{1}{c|}{\textbf{0.963}}  & \multicolumn{1}{c|}{\textbf{0.262}}  & \multicolumn{1}{c|}{0.265}  & \multicolumn{1}{c|}{\textbf{0.972}}  & \multicolumn{1}{c|}{\textbf{0.967}}  & \multicolumn{1}{c|}{\textbf{0.913}}  & \multicolumn{1}{c||}{\textbf{0.887}}  & \multicolumn{1}{c|}{0.968*}  & \multicolumn{1}{c|}{0.968*}  & \multicolumn{1}{c|}{0.518}  & \multicolumn{1}{c|}{0.528}
\\ \hline
\multicolumn{1}{|l||}{\textbf{LoST-X}}  & \multicolumn{1}{c|}{0.046}  & \multicolumn{1}{c|}{0.506}  & \multicolumn{1}{c|}{0.037}  & \multicolumn{1}{c|}{0.341}  & \multicolumn{1}{c|}{0.025}  & \multicolumn{1}{c|}{0.169}  & \multicolumn{1}{c|}{0.057}  & \multicolumn{1}{c|}{0.451}  & \multicolumn{1}{c|}{0.049}  & \multicolumn{1}{c||}{0.392}  & \multicolumn{1}{c|}{\textbf{0.999}}  & \multicolumn{1}{c|}{\textbf{1.000}}  & \multicolumn{1}{c|}{0.566}  & \multicolumn{1}{c|}{0.627}
\\ \hline
\multicolumn{1}{|l||}{\textbf{LoST+SM}}  & \multicolumn{1}{c|}{0.027}  & \multicolumn{1}{c|}{0.864}  & \multicolumn{1}{c|}{0.006}  & \multicolumn{1}{c|}{0.657}  & \multicolumn{1}{c|}{0.022}  & \multicolumn{1}{c|}{0.259}  & \multicolumn{1}{c|}{0.086}  & \multicolumn{1}{c|}{0.861}  & \multicolumn{1}{c|}{0.074}  & \multicolumn{1}{c||}{0.752}  & \multicolumn{1}{c|}{0.978*}  & \multicolumn{1}{c|}{0.978*}  & \multicolumn{1}{c|}{\textbf{0.837}}  & \multicolumn{1}{c|}{\textbf{0.844}}
\\ \hline
\multicolumn{1}{|l||}{\textbf{LoST-X+SM}}  & \multicolumn{1}{c|}{0.019}  & \multicolumn{1}{c|}{0.859}  & \multicolumn{1}{c|}{0.004}  & \multicolumn{1}{c|}{0.640}  & \multicolumn{1}{c|}{0.019}  & \multicolumn{1}{c|}{\textbf{0.296}}  & \multicolumn{1}{c|}{0.062}  & \multicolumn{1}{c|}{0.873}  & \multicolumn{1}{c|}{0.050}  & \multicolumn{1}{c||}{0.795}  & \multicolumn{1}{c|}{0.978*}  & \multicolumn{1}{c|}{0.978*}  & \multicolumn{1}{c|}{0.808}  & \multicolumn{1}{c|}{0.835}
\\ \hline
\multicolumn{1}{|l||}{\textbf{Seq2single}}  & \multicolumn{1}{c|}{0.032}  & \multicolumn{1}{c|}{0.545}  & \multicolumn{1}{c|}{0.025}  & \multicolumn{1}{c|}{0.289}  & \multicolumn{1}{c|}{0.015}  & \multicolumn{1}{c|}{0.112}  & \multicolumn{1}{c|}{0.049}  & \multicolumn{1}{c|}{0.471}  & \multicolumn{1}{c|}{0.034}  & \multicolumn{1}{c||}{0.379}  & \multicolumn{1}{c|}{0.990}  & \multicolumn{1}{c|}{0.991}  & \multicolumn{1}{c|}{0.338}  & \multicolumn{1}{c|}{0.379}
\\ \hline
\multicolumn{1}{|l||}{\textbf{NetVLAD+SM}}  & \multicolumn{1}{c|}{0.001}  & \multicolumn{1}{c|}{0.874}  & \multicolumn{1}{c|}{0.000}  & \multicolumn{1}{c|}{0.361}  & \multicolumn{1}{c|}{0.000}  & \multicolumn{1}{c|}{0.112}  & \multicolumn{1}{c|}{0.014}  & \multicolumn{1}{c|}{0.850}  & \multicolumn{1}{c|}{0.003}  & \multicolumn{1}{c||}{0.791}  & \multicolumn{1}{c|}{0.978*}  & \multicolumn{1}{c|}{0.978*}  & \multicolumn{1}{c|}{0.666}  & \multicolumn{1}{c|}{0.672}
\\ \hline
\multicolumn{1}{|l||}{\textbf{SO-DSO VPR}}  & \multicolumn{1}{c|}{0.294}  & \multicolumn{1}{c|}{0.444}  & \multicolumn{1}{c|}{0.225}  & \multicolumn{1}{c|}{0.331}  & \multicolumn{1}{c|}{0.018}  & \multicolumn{1}{c|}{0.047}  & \multicolumn{1}{c|}{0.291}  & \multicolumn{1}{c|}{0.442}  & \multicolumn{1}{c|}{0.150}  & \multicolumn{1}{c||}{0.278}  & \multicolumn{1}{c|}{0.946}  & \multicolumn{1}{c|}{0.948}  & \multicolumn{1}{c|}{0.101}  & \multicolumn{1}{c|}{0.117}
\\ \hline
\multicolumn{15}{l}{\textbf{15 m / 80 m}: 15 and 80 meter localization radii, *: zero incorrect matches}
\end{tabular}
}
\label{table:pr_performance}
\vspace{-4mm}
\end{table*}

The MR100 and AUC values achieved by each method with each query sequence are presented in Table \ref{table:pr_performance}. With the opposing viewpoint query sequences, SPOT demonstrates remarkable improvement over the baselines, achieving 91.7\% MR100 with the \textit{Noon R0} sequence and 83.6\% MR100 with the dimly lit \textit{Sunset R0} under the strict 15 meter localization radius. Performance is also strong on the \textit{Diverge R1} sequence, which shows SPOT can reliably identify true negatives. The opposing viewpoint \textit{Night R0} sequence is more challenging, with successful matches limited primarily to a well-lit urban portion. Altogether, the results indicate that SPOT can achieve excellent performance under opposing viewpoints and changes in lighting conditions so long as surrounding structure can still be perceived. Note that in some cases, there is even \textit{worse} performance with the 80 meter localization radius because some of the true negatives with the 15 meter radius become false negatives with the 80 meter radius.

\begin{figure}
\centering
\includegraphics[width=1.0\columnwidth]{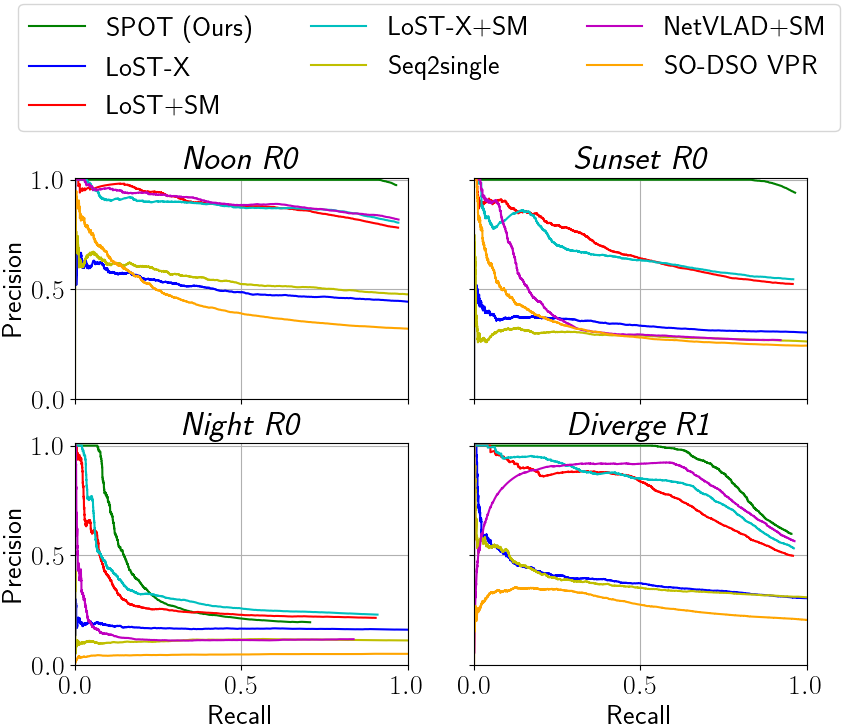}
\caption{Precision-recall curves for selected opposing viewpoint query sequences using an 80 meter localization radius.}  
\label{figure:pr_curves}
\vspace{-6mm}
\end{figure}

In contrast to the proposed method, the baselines achieve at most 10\% MR100 under opposing viewpoints with the 80 meter localization radius. The AUC results and full PR curves (Fig.~\ref{figure:pr_curves}), indicate the baselines do generate a meaningful number of true positive matches within the 80 meter localization radius. However, with the stricter 15 meter localization radius, only the structure based SO-DSO VPR attains an AUC greater than 0.09 under opposing viewpoints. This localization inaccuracy is likely due to the \textit{visual offset} problem described in Section~\ref{section:opposing_viewpoint_vpr}.

With similar viewpoints, all methods perform well under the 15 meter localization threshold. In terms of MR100, methods with sequence matching perform better. SPOT performs competitively on \textit{Night R0}, although it is surpassed by several of the baselines on this sequence.

\subsection{Computation Time and Storage Requirements}

\begin{table}[]
\centering
\caption{Computation time per query (opposing \textit{Noon R0}) and storage}
\resizebox{1.00\columnwidth}{!}{
\begin{tabular}{cccccc}
\hline
\multicolumn{1}{|c||}{\multirow{2}{*}{\textbf{Method}}}  & \multicolumn{2}{c|}{\begin{tabular}[c]{@{}c@{}}\textbf{Query}\\\textbf{Place Description}\end{tabular}}  & \multicolumn{2}{c|}{\textbf{Matching}} & \multicolumn{1}{|l|}{\multirow{3}{*}{\textbf{\begin{tabular}[c]{@{}l@{}}Storage\\ Per\\ Reference\end{tabular}}}}
\\ \cline{2-5}
\multicolumn{1}{|l||}{}  & \multicolumn{1}{c|}{\begin{tabular}[c]{@{}c@{}}\textbf{Average}\\\textbf{Time (ms)}\end{tabular}}  & \multicolumn{1}{c|}{\textbf{HW}}  & \multicolumn{1}{c|}{\begin{tabular}[c]{@{}c@{}}\textbf{Average}\\\textbf{Time (ms)}\end{tabular}}  & \multicolumn{1}{c|}{\textbf{HW}} & \multicolumn{1}{|l|}{}
\\ \hline\hline
\multicolumn{1}{|l||}{\textbf{SPOT (Ours)}}  & \multicolumn{1}{c|}{\textbf{0.16}}  & \multicolumn{1}{c|}{C-1}  & \multicolumn{1}{c|}{11.86}  & \multicolumn{1}{c|}{C-M} & \multicolumn{1}{c|}{\textbf{5.0 KB}}
\\ \hline
\multicolumn{1}{|l||}{\textbf{LoST-X}}  & \multicolumn{1}{c|}{131.17}  & \multicolumn{1}{c|}{G/C-1}  & \multicolumn{1}{c|}{85.03}  & \multicolumn{1}{c|}{C-1} & \multicolumn{1}{c|}{15.8 MB}
\\ \hline
\multicolumn{1}{|l||}{\textbf{LoST+SM}}  & \multicolumn{1}{c|}{131.17}  & \multicolumn{1}{c|}{G/C-1}  & \multicolumn{1}{c|}{21.64}  & \multicolumn{1}{c|}{C-1/M} & \multicolumn{1}{c|}{49.2 KB}
\\ \hline
\multicolumn{1}{|l||}{\textbf{LoST-X+SM}}  & \multicolumn{1}{c|}{131.17}  & \multicolumn{1}{c|}{G/C-1}  & \multicolumn{1}{c|}{93.06}  & \multicolumn{1}{c|}{C-1/M} & \multicolumn{1}{c|}{15.8 MB}
\\ \hline
\multicolumn{1}{|l||}{\textbf{Seq2single}}  & \multicolumn{1}{c|}{128.39}  & \multicolumn{1}{c|}{G}  & \multicolumn{1}{c|}{1962.16}  & \multicolumn{1}{c|}{C-1} & \multicolumn{1}{c|}{16.3 MB}
\\ \hline
\multicolumn{1}{|l||}{\textbf{NetVLAD+SM}}  & \multicolumn{1}{c|}{32.52}  & \multicolumn{1}{c|}{G}  & \multicolumn{1}{c|}{12.23}  & \multicolumn{1}{c|}{C-1/M} & \multicolumn{1}{c|}{16.4 KB}
\\ \hline
\multicolumn{1}{|l||}{\textbf{SO-DSO VPR}}  & \multicolumn{1}{c|}{0.66}  & \multicolumn{1}{c|}{C-1}  & \multicolumn{1}{c|}{\textbf{5.39}}  & \multicolumn{1}{c|}{C-M} & \multicolumn{1}{c|}{19.2 KB}
\\ \hline
\multicolumn{6}{l}{\textbf{HW}: Hardware, \textbf{G}: GPU process, \textbf{C-1/M}: single/multi-threaded CPU process}
\end{tabular}
}
\label{table:computation_time_and_storage}
\vspace{-4mm}
\end{table}

Table \ref{table:computation_time_and_storage} lists computation time and reference storage space requirements for each method. All experiments were run on a machine with an AMD Ryzen 9 5950x 16-core, 32-thread CPU and a NVIDIA RTX A6000 GPU. The hardware utilized by each stage of each method is specified in the table. SPOT is more efficient in both aspects than nearly all baselines tested. LoST-X and Seq2single require the high dimensional output of an intermediate convolutional layer to be retained for each reference image resulting in storage requirements three orders of magnitude greater than all other methods tested. 

\section{Limitations}\label{section:limitations}
Several assumptions are implicit in our method that are frequently valid in driving scenarios but not generally applicable. Specifically, the method assumes: (1) predominantly straight motion, (2) similar or opposing viewpoint (yaw variation of 0 or 180 degrees), (3) limited roll-pitch rotations, and (4) a known, constant camera height above the ground. The current implementation also assumes forward-facing cameras, so modifications would be required to apply this method to configurations where the cameras oppose the direction of travel.

\section{Conclusions \& Future Work}
In this paper, we presented SPOT, a technique for opposing viewpoint VPR that relies exclusively on structure estimated through stereo VO. Evaluating SPOT against several baselines on the NSAVP dataset, we demonstrated remarkable improvement over the state-of-the-art. Overall, we believe SPOT further signals the potential of VO-derived structure for VPR. The relatively low localization error of our approach makes it promising for future integration within a SLAM system to support opposing viewpoint loop closure or multi-map merging. Moreover, future work could assess the proposed method's suitability with monocular visual inertial odometry and for cross-modality place recognition, using point clouds as a common representation.

%\addtolength{\textheight}{-12cm}   % This command serves to balance the column lengths
                                  % on the last page of the document manually. It shortens
                                  % the textheight of the last page by a suitable amount.
                                  % This command does not take effect until the next page
                                  % so it should come on the page before the last. Make
                                  % sure that you do not shorten the textheight too much.

%%%%%%%%%%%%%%%%%%%%%%%%%%%%%%%%%%%%%%%%%%%%%%%%%%%%%%%%%%%%%%%%%%%%%%%%%%%%%%%%

%%%%%%%%%%%%%%%%%%%%%%%%%%%%%%%%%%%%%%%%%%%%%%%%%%%%%%%%%%%%%%%%%%%%%%%%%%%%%%%%

%%%%%%%%%%%%%%%%%%%%%%%%%%%%%%%%%%%%%%%%%%%%%%%%%%%%%%%%%%%%%%%%%%%%%%%%%%%%%%%%

% \section*{Acknowledgment}

% This work was supported by a grant from Ford Motor Company via the Ford-UM Alliance under award N028603.

\ifthenelse{\boolean{preprint}}{\begin{appendices}

\addtocounter{table}{1}
\begin{table*}
\centering
\caption{Ablation study results using a 15 meter localization radius}
\resizebox{1.00\textwidth}{!}{
\begin{tabular}{cccccccccccccccc}
\hline
\multicolumn{1}{|c||}{\multirow{3}{*}{\textbf{Method}}}  & \multicolumn{10}{c||}{\textbf{Opposing Viewpoint}}  & \multicolumn{4}{c||}{\textbf{Similar Viewpoint}}  & \multicolumn{1}{c|}{\multirow{3}{*}{\textbf{MT}}} 
\\ \cline{2-15}
\multicolumn{1}{|l||}{}  & \multicolumn{2}{c|}{\textit{\textbf{Noon R0}}}  & \multicolumn{2}{c|}{\textit{\textbf{Sunset R0}}}  & \multicolumn{2}{c|}{\textit{\textbf{Night R0}}}  & \multicolumn{2}{c|}{\textit{\textbf{Noon R1}}}  & \multicolumn{2}{c||}{\textit{\textbf{Diverge R1}}}  & \multicolumn{2}{c|}{\textit{\textbf{Sunset R0}}}  & \multicolumn{2}{c||}{\textit{\textbf{Night R0}}}  & \multicolumn{1}{c|}{} 
\\ \cline{2-15}
\multicolumn{1}{|l||}{}  & \multicolumn{1}{c|}{\textbf{MR100}}  & \multicolumn{1}{c|}{\textbf{AUC}}  & \multicolumn{1}{c|}{\textbf{MR100}}  & \multicolumn{1}{c|}{\textbf{AUC}}  & \multicolumn{1}{c|}{\textbf{MR100}}  & \multicolumn{1}{c|}{\textbf{AUC}}  & \multicolumn{1}{c|}{\textbf{MR100}}  & \multicolumn{1}{c|}{\textbf{AUC}}  & \multicolumn{1}{c|}{\textbf{MR100}}  & \multicolumn{1}{c||}{\textbf{AUC}}  & \multicolumn{1}{c|}{\textbf{MR100}}  & \multicolumn{1}{c|}{\textbf{AUC}}  & \multicolumn{1}{c|}{\textbf{MR100}}  & \multicolumn{1}{c||}{\textbf{AUC}}  & \multicolumn{1}{c|}{} 
\\ \hline\hline
\multicolumn{1}{|l||}{\textbf{A-CC}}  & \multicolumn{1}{c|}{0.007}  & \multicolumn{1}{c|}{0.089}  & \multicolumn{1}{c|}{0.002}  & \multicolumn{1}{c|}{0.072}  & \multicolumn{1}{c|}{0.000}  & \multicolumn{1}{c|}{0.007}  & \multicolumn{1}{c|}{0.001}  & \multicolumn{1}{c|}{0.073}  & \multicolumn{1}{c|}{0.001}  & \multicolumn{1}{c||}{0.038}  & \multicolumn{1}{c|}{0.401}  & \multicolumn{1}{c|}{0.794}  & \multicolumn{1}{c|}{0.002}  & \multicolumn{1}{c||}{0.010}  & \multicolumn{1}{c|}{\textbf{0.08}} 
\\ \hline
\multicolumn{1}{|l||}{\textbf{A-CC+NN}}  & \multicolumn{1}{c|}{0.014}  & \multicolumn{1}{c|}{0.448}  & \multicolumn{1}{c|}{0.008}  & \multicolumn{1}{c|}{0.303}  & \multicolumn{1}{c|}{0.000}  & \multicolumn{1}{c|}{0.010}  & \multicolumn{1}{c|}{0.000}  & \multicolumn{1}{c|}{0.211}  & \multicolumn{1}{c|}{0.001}  & \multicolumn{1}{c||}{0.122}  & \multicolumn{1}{c|}{0.509}  & \multicolumn{1}{c|}{0.992}  & \multicolumn{1}{c|}{0.000}  & \multicolumn{1}{c||}{0.136}  & \multicolumn{1}{c|}{0.45} 
\\ \hline
\multicolumn{1}{|l||}{\textbf{A-CC+CD+NN}}  & \multicolumn{1}{c|}{0.003}  & \multicolumn{1}{c|}{0.536}  & \multicolumn{1}{c|}{0.000}  & \multicolumn{1}{c|}{0.380}  & \multicolumn{1}{c|}{0.000}  & \multicolumn{1}{c|}{0.031}  & \multicolumn{1}{c|}{0.003}  & \multicolumn{1}{c|}{0.493}  & \multicolumn{1}{c|}{0.006}  & \multicolumn{1}{c||}{0.305}  & \multicolumn{1}{c|}{0.515}  & \multicolumn{1}{c|}{\textbf{0.993}}  & \multicolumn{1}{c|}{0.005}  & \multicolumn{1}{c||}{0.154}  & \multicolumn{1}{c|}{0.44} 
\\ \hline
\multicolumn{1}{|l||}{\textbf{A-CC+CD+SM}}  & \multicolumn{1}{c|}{0.708}  & \multicolumn{1}{c|}{0.955}  & \multicolumn{1}{c|}{0.621}  & \multicolumn{1}{c|}{0.948}  & \multicolumn{1}{c|}{0.028}  & \multicolumn{1}{c|}{0.145}  & \multicolumn{1}{c|}{0.702}  & \multicolumn{1}{c|}{0.960}  & \multicolumn{1}{c|}{0.382}  & \multicolumn{1}{c||}{0.853}  & \multicolumn{1}{c|}{0.966}  & \multicolumn{1}{c|}{0.968}  & \multicolumn{1}{c|}{0.156}  & \multicolumn{1}{c||}{0.407}  & \multicolumn{1}{c|}{2.63} 
\\ \hline
\multicolumn{1}{|l||}{\textbf{SPOT}}  & \multicolumn{1}{c|}{\textbf{0.917}}  & \multicolumn{1}{c|}{\textbf{0.970}}  & \multicolumn{1}{c|}{\textbf{0.836}}  & \multicolumn{1}{c|}{\textbf{0.968}}  & \multicolumn{1}{c|}{\textbf{0.067}}  & \multicolumn{1}{c|}{\textbf{0.262}}  & \multicolumn{1}{c|}{\textbf{0.798}}  & \multicolumn{1}{c|}{\textbf{0.972}}  & \multicolumn{1}{c|}{\textbf{0.605}}  & \multicolumn{1}{c||}{\textbf{0.913}}  & \multicolumn{1}{c|}{\textbf{0.968}*}  & \multicolumn{1}{c|}{0.968*}  & \multicolumn{1}{c|}{\textbf{0.199}}  & \multicolumn{1}{c||}{\textbf{0.518}}  & \multicolumn{1}{c|}{11.86} 
\\ \hline
\multicolumn{16}{l}{*: zero incorrect matches, \textbf{MR100}: maximum recall at 100\% precision, \textbf{MT}: average matching computation time per query (in ms) with the \textit{Noon R0} query sequence}
\end{tabular}
}
\label{table:ablation_study_results}
\vspace{-4mm}
\end{table*}

\section{Ablation Study}\label{appendix:ablation_study}

\addtocounter{table}{-2}
\begin{table}[]
\centering
\caption{Methods evaluated in the ablation study.}
\resizebox{\columnwidth}{!}{%
\begin{tabular}{|l||lllllll|}
\hline
\multicolumn{1}{|c||}{{\textbf{\begin{tabular}[c]{@{}c@{}}Place\\ Recognition\\ Method\end{tabular}}}} & \multicolumn{3}{c|}{\textbf{\begin{tabular}[c]{@{}c@{}}Descriptor\\ Distance\\ Method\end{tabular}}}                              & \multicolumn{4}{c|}{\textbf{\begin{tabular}[c]{@{}c@{}}Matching\\ Method\end{tabular}}}                         \\ \cline{2-8} 
                                                                                               & \multicolumn{1}{l|}{\textbf{SC}}         & \multicolumn{1}{l|}{\textbf{CD}}         & \multicolumn{1}{l|}{\textbf{VD}}         & \multicolumn{1}{l|}{\textbf{RK}}           & \multicolumn{1}{l|}{\textbf{NN}}    & \multicolumn{1}{l|}{\textbf{SM}}       & \textbf{DD}           \\ \hline
\textbf{A-CC \cite{kim_scan_2022}}                                                                                  &  \multicolumn{1}{c|}{$\times$}             & \multicolumn{1}{l|}{}                    & \multicolumn{1}{l|}{}                    & \multicolumn{1}{c|}{$\times$}              & \multicolumn{1}{l|}{}     &        \multicolumn{1}{l|}{}          &                       \\ \hline
\textbf{A-CC+NN}                                                                               & \multicolumn{1}{c|}{$\times$}             & \multicolumn{1}{l|}{}                    & \multicolumn{1}{l|}{}                    & \multicolumn{1}{l|}{}                      & \multicolumn{1}{c|}{$\times$}              & \multicolumn{1}{l|}{}     &                  \\ \hline
\textbf{A-CC+CD+NN}                                                                            &  \multicolumn{1}{l|}{}                     & \multicolumn{1}{c|}{$\times$}            & \multicolumn{1}{l|}{}                    & \multicolumn{1}{l|}{}                      & \multicolumn{1}{c|}{$\times$}              &  \multicolumn{1}{l|}{}    &                  \\ \hline
\textbf{A-CC+CD+SM}                                                                            &  \multicolumn{1}{l|}{}                     & \multicolumn{1}{c|}{$\times$}            & \multicolumn{1}{l|}{}                    & \multicolumn{1}{l|}{}                      & \multicolumn{1}{l|}{}                      & \multicolumn{1}{c|}{$\times$} &             \\ \hline
\textbf{SPOT}                                                                                &  \multicolumn{1}{l|}{}                     & \multicolumn{1}{l|}{}                    & \multicolumn{1}{c|}{$\times$}            & \multicolumn{1}{l|}{}                      & \multicolumn{1}{l|}{}          &     \multicolumn{1}{l|}{}        & \multicolumn{1}{c|}{$\times$}              \\ \hline\hline
\multicolumn{1}{|c||}{\textbf{\begin{tabular}[c]{@{}c@{}}Descriptor\\ Distance Method\end{tabular}}}                   & \multicolumn{7}{c|}{\textbf{Description}}                                                                                                                                                                                                                                                                                                                                  \\ \hline
\textbf{SC}                                                                                   & \multicolumn{7}{l|}{\begin{tabular}[c]{@{}l@{}}Query and reference are aligned with\\ an aligning key and the column-wise\\ cosine distance is computed between them~\cite{kim_scan_2022}\end{tabular}}                                                                                                                                                                     \\ \hline
\textbf{CD}                                                                                    & \multicolumn{7}{l|}{\begin{tabular}[c]{@{}l@{}}The column-wise cosine distance in SC is \\ replaced with the cosine distance \end{tabular}}                                                                                                                                     \\ \hline
\textbf{VD}                                                                                    & \multicolumn{7}{l|}{\begin{tabular}[c]{@{}l@{}}Variable offset double descriptor distance as \\described in Section~\ref{section:variable_offset_double_descriptor_distance_computation}\end{tabular}}                                                                                                                                                      \\ \hline\hline
\multicolumn{1}{|c||}{\textbf{\begin{tabular}[c]{@{}c@{}}Matching \\ Method\end{tabular}}}                                                                          & \multicolumn{7}{c|}{\textbf{Description}}                                                                                                                                                                                                                                                                                                                                  \\ \hline
\textbf{RK}                                                                                    & \multicolumn{7}{l|}{\begin{tabular}[c]{@{}l@{}}Matching is performed with a retrieval \\key KD tree~\cite{kim_scan_2022}\end{tabular}}                                                                                                                                                                                                                                                                                \\ \hline
\textbf{NN}                                                                                    & \multicolumn{7}{l|}{Nearest neighbor matching}                                                                                                                                                                                                                                                                                                                 \\ \hline
\textbf{SM}                                                                                    & \multicolumn{7}{l|}{\begin{tabular}[c]{@{}l@{}}Sequence matching with both positive and\\ negative slopes evaluated in a single distance \\matrix which is the element-wise minimum of \\$D_{sim}$ and $D_{opp}$ \end{tabular}} \\ \hline
\textbf{DD}                                                                                    & \multicolumn{7}{l|}{\begin{tabular}[c]{@{}l@{}}Double distance matrix sequence matching as \\described in Section~\ref{section:double_distance_matrix_sequence_matching}\end{tabular}} \\ \hline
\end{tabular}%
}
\label{table:ablation_study_methods}
\vspace{-4mm}
\end{table}

To assess the impact of various components of our method, we evaluate several variants. First, we directly apply the A-CC place recognition method \cite{kim_scan_2022} to the keyframe point clouds obtained from accumulating the output of SO-DSO \cite{mo_extending_2019}. Then, we evaluate progressive modifications to the descriptor distance computation and matching components of the A-CC method that bring it closer to SPOT. Table~\ref{table:ablation_study_methods} details the variants tested. The parameter values from Table~\ref{table:parameters} are used for all methods, where applicable. We implemented all methods within a single, modular framework.

The ablation study results are presented in Table \ref{table:ablation_study_results}. These results show that direct application of the A-CC method \cite{kim_scan_2022} to the keyframe point clouds yields poor performance in opposing viewpoint cases. A-CC+NN and A-CC+CD+NN show significant improvements in terms of AUC, but not MR100. The addition of sequence matching substantially improves the MR100 in almost all cases, as illustrated by the difference between A-CC+CD+NN and A-CC+CD+SM. The variable offset descriptor distance and double distance matrix sequence matching methods introduced in SPOT produce another significant boost to the MR100 achieved in most cases, although it comes at the cost of increased matching computation time. 

\begin{figure}
\centering
\includegraphics[width=\columnwidth]{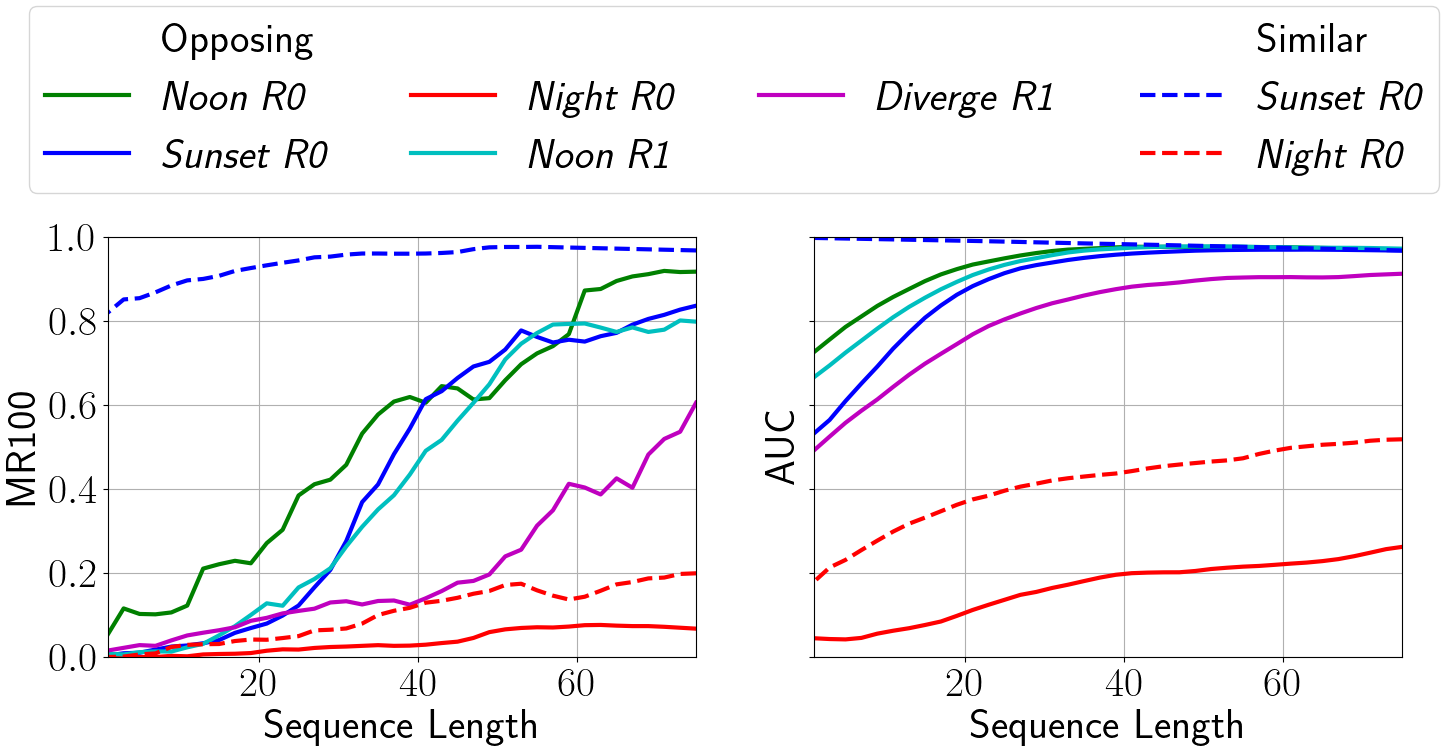}
\caption{SPOT's performance as a function of sequence length using a 15 meter localization radius. \vspace{-4mm}}
\label{figure:sequence_study_results}
\end{figure}

\section{Sequence Study}\label{appendix:sequence_study}

Sequence matching imposes a latency between first entering a revisited area and recognizing a place, which scales with the sequence length $w$. In addition, predicted matches are given for the query at the center of the search window, meaning that there is a $(w - 1) / 2$ descriptor offset between a recognized place and the current camera pose. Moreover, matching computation time increases with $w$. For these reasons, it is advantageous to minimize $w$. We performed a study of SPOT's performance as a function of $w$.

The results of this study are displayed in Fig.~\ref{figure:sequence_study_results}. For all but the \textit{Night} query sequences, the AUC remains substantial ($\geq$ 0.49) regardless of $w$. On the other hand, the maximum recall at 100\% precision of opposing viewpoint query sequences consistently improves with increasing $w$. Taken together, this presents a tradeoff: a small $w$ can be used if the application is not sensitive to false positives or includes a means of rejecting them, or a large $w$ can be used to avoid false positives. The total matching computation time only grows moderately with increasing $w$. With $w=1$, the total matching computation time is $\sim$9.3 ms and $\sim$11.2 ms for the \textit{R0} and \textit{R1} sequences, respectively, and with $w=75$, these times are only $\sim$2 ms greater. The impact of $w$ on the total matching computation time is relatively small because the total matching computation time is dominated by descriptor distance calculation, the results of which are stored and reused as the sequence search window slides over the columns of the distance matrices. 

\end{appendices}}{}

\bibliographystyle{IEEEtran}
\bibliography{IEEEabrv, spot}

@misc{projectpage,
  howpublished = {{Project} webpage: https://umautobots.github.io/spot}
}

@inproceedings{mo_extending_2019,
	address = {Macau, China},
    title={Extending Monocular Visual Odometry to Stereo Camera Systems by Scale optimization},
	booktitle={2019 IEEE/RSJ International Conference on Intelligent Robots and Systems (IROS)},
	author={Mo, Jiawei and Sattar, Junaed},
	month = nov,
	year = {2019},
	pages={6921-6927},
}

@ARTICLE{kim_scan_2022,
  author={Kim, Giseop and Choi, Sunwook and Kim, Ayoung},
  journal={IEEE Transactions on Robotics}, 
  title={Scan Context++: Structural Place Recognition Robust to Rotation and Lateral Variations in Urban Environments},
  month = jun,
  year={2022},
  volume={38},
  number={3},
  pages={1856-1874},
}

@ARTICLE{campos_orb-slam3_2021,
  author={Campos, Carlos and Elvira, Richard and Rodríguez, Juan J. Gómez and M. Montiel, José M. and D. Tardós, Juan},
  journal={IEEE Transactions on Robotics}, 
  title={{ORB}-{SLAM3}: An Accurate Open-Source Library for Visual, Visual–Inertial, and Multimap {SLAM}}, 
  month = dec,
  year={2021},
  volume={37},
  number={6},
  pages={1874-1890},
}

@ARTICLE{lowry_visual_2016,
  author={Lowry, Stephanie and Sünderhauf, Niko and Newman, Paul and Leonard, John J. and Cox, David and Corke, Peter and Milford, Michael J.},
  journal={IEEE Transactions on Robotics}, 
  title={Visual Place Recognition: A Survey}, 
  month = feb,
  year={2016},
  volume={32},
  number={1},
  pages={1-19},
}

@article{zhang_visual_2021,
  title={Visual place recognition: A survey from deep learning perspective},
  author={Zhang, Xiwu and Wang, Lei and Su, Yan},
  journal={Pattern Recognition},
  volume={113},
  pages={107760},
  month = may,
  year={2021},
}

@INPROCEEDINGS{arandjelovic_netvlad_2016,
  address={Las Vegas, NV, USA},
  author={Arandjelovic, Relja and Gronat, Petr and Torii, Akihiko and Pajdla, Tomas and Sivic, Josef},
  booktitle={2016 IEEE Conference on Computer Vision and Pattern Recognition (CVPR)}, 
  title={{NetVLAD}: {CNN} Architecture for Weakly Supervised Place Recognition}, 
  month = jun,
  year={2016},
  pages={5297-5307},
}

@INPROCEEDINGS{ali2023mixvpr,
  address={Waikoloa, HI, USA},
  author={Ali-Bey, Amar and Chaib-Draa, Brahim and Giguére, Philippe},
  booktitle={2023 IEEE/CVF Winter Conference on Applications of Computer Vision (WACV)}, 
  title={{MixVPR}: Feature Mixing for Visual Place Recognition}, 
  month=jan,
  year={2023},
  pages={2997-3006},
}

@ARTICLE{keetha2023anyloc,
  author={Keetha, Nikhil and Mishra, Avneesh and Karhade, Jay and Jatavallabhula, Krishna Murthy and Scherer, Sebastian and Krishna, Madhava and Garg, Sourav},
  journal={IEEE Robotics and Automation Letters}, 
  title={{AnyLoc}: Towards Universal Visual Place Recognition}, 
  month=feb,
  year={2024},
  volume={9},
  number={2},
  pages={1286-1293},
}

@ARTICLE{gawel_x-view_2018,
  author={Gawel, Abel and Don, Carlo Del and Siegwart, Roland and Nieto, Juan and Cadena, Cesar},
  journal={IEEE Robotics and Automation Letters}, 
  title={{X}-{View}: Graph-Based Semantic Multi-View Localization}, 
  month = jul,
  year={2018},
  volume={3},
  number={3},
  pages={1687-1694},
}

@INPROCEEDINGS{garg_dont_2018,
  address={Brisbane, QLD, Australia},
  author={Garg, Sourav and Suenderhauf, Niko and Milford, Michael},
  booktitle={2018 IEEE International Conference on Robotics and Automation (ICRA)}, 
  title={Don't Look Back: Robustifying Place Categorization for Viewpoint- and Condition-Invariant Place Recognition},
  month = may,
  year={2018},
  pages={3645-3652},
}

@INPROCEEDINGS{garg_lost_2018, 
    author    = {Sourav Garg AND Niko Suenderhauf AND Michael Milford}, 
    title     = {{LoST}? Appearance-Invariant Place Recognition for Opposite Viewpoints using Visual Semantics}, 
    booktitle = {Proceedings of Robotics: Science and Systems}, 
    year      = {2018}, 
    address   = {Pittsburgh, PA, USA}, 
    month     = jun, 
    pages={1-10},
}

@INPROCEEDINGS{garg_look_2019,
  address={Montreal, QC, Canada},
  author={Garg, Sourav and Babu V, Madhu and Dharmasiri, Thanuja and Hausler, Stephen and Suenderhauf, Niko and Kumar, Swagat and Drummond, Tom and Milford, Michael},
  booktitle={2019 International Conference on Robotics and Automation (ICRA)}, 
  title={Look No Deeper: Recognizing Places from Opposing Viewpoints under Varying Scene Appearance using Single-View Depth Estimation}, 
  month = may,
  year={2019},
  pages={4916-4923},
}

@article{garg_semanticgeometric_2022,
  title={Semantic--geometric visual place recognition: a new perspective for reconciling opposing views},
  author={Garg, Sourav and Suenderhauf, Niko and Milford, Michael},
  journal={The International Journal of Robotics Research},
  volume={41},
  number={6},
  pages={573--598},
  month=may,
  year={2022},
}

@INPROCEEDINGS{arroyo_bidirectional_2014,
  address={Dearborn, MI, USA},
  author={Arroyo, Roberto and Alcantarilla, Pablo F. and Bergasa, Luis M. and Yebes, J. Javier and Gámez, Sergio},
  booktitle={2014 IEEE Intelligent Vehicles Symposium Proceedings}, 
  title={Bidirectional loop closure detection on panoramas for visual navigation}, 
  month = jun,
  year={2014},
  pages={1378-1383},
}

@INPROCEEDINGS{multisphere,
  address={Barcelona, Spain},
  author={Chapoulie, Alexandre and Rives, Patrick and Filliat, David},
  booktitle={2011 IEEE International Conference on Computer Vision Workshops (ICCV Workshops)}, 
  title={A spherical representation for efficient visual loop closing}, 
  month=nov,
  year={2011},
  pages={335-342},
}

@article{xu2023leveraging,
  title={Leveraging {BEV} Representation for 360-degree Visual Place Recognition},
  author={Xu, Xuecheng and Jiao, Yanmei and Lu, Sha and Ding, Xiaqing and Xiong, Rong and Wang, Yue},
  journal={arXiv preprint arXiv:2305.13814},
  month=may,
  year={2023},
  pages={1-11}
}

@INPROCEEDINGS{cieslewski_point_2016,
  address={Stockholm, Sweden},
  author={Cieslewski, Titus and Stumm, Elena and Gawel, Abel and Bosse, Mike and Lynen, Simon and Siegwart, Roland},
  booktitle={2016 IEEE International Conference on Robotics and Automation (ICRA)}, 
  title={Point cloud descriptors for place recognition using sparse visual information},
  month=may,
  year={2016},
  pages={4830-4836},
}

@inproceedings{ye_place_2017,
  address={London, UK},
  author = {Ye, Yawei and Cieslewski, Titus and Loquercio, Antonio and Scaramuzza, Davide},
  title = {Place recognition in semi-dense maps: Geometric and learning-based approaches},
  booktitle    = {British Machine Vision Conference},
  month = sep,
  year         = {2017},
  pages = {1--13},
}

@INPROCEEDINGS{mo_fast_2020,
  address={Las Vegas, NV, USA},
  author={Mo, Jiawei and Sattar, Junaed},
  booktitle={2020 IEEE/RSJ International Conference on Intelligent Robots and Systems (IROS)}, 
  title={A Fast and Robust Place Recognition Approach for Stereo Visual Odometry Using {LiDAR} Descriptors}, 
  month=oct,
  year={2020},
  pages={5893-5900},
}

@ARTICLE{oertel_augmenting_2020,
  author={Oertel, Amadeus and Cieslewski, Titus and Scaramuzza, Davide},
  journal={IEEE Robotics and Automation Letters}, 
  title={Augmenting Visual Place Recognition With Structural Cues}, 
  month=oct,
  year={2020},
  volume={5},
  number={4},
  pages={5534-5541},
}

@article{nsavp,
author = {Spencer Carmichael and Austin Buchan and Mani Ramanagopal and Radhika Ravi and Ram Vasudevan and Katherine A Skinner},
title = {Dataset and Benchmark: Novel Sensors for Autonomous Vehicle Perception},
journal = {The International Journal of Robotics Research},
volume = {44},
number = {3},
pages = {355-365},
year = {2025},
}

@INPROCEEDINGS{jegou_aggregating_2010,
  address={San Francisco, CA, USA},
  author={Jégou, Hervé and Douze, Matthijs and Schmid, Cordelia and Pérez, Patrick},
  booktitle={2010 IEEE Computer Society Conference on Computer Vision and Pattern Recognition}, 
  title={Aggregating local descriptors into a compact image representation}, 
  month=jun,
  year={2010},
  pages={3304-3311},
}

@ARTICLE{masone_survey_2021,
  author={Masone, Carlo and Caputo, Barbara},
  journal={IEEE Access}, 
  title={A Survey on Deep Visual Place Recognition}, 
  month=jan,
  year={2021},
  volume={9},
  pages={19516-19547},
}

@inproceedings{zaffar_levelling_2019,
    address={Montreal, Canada},
	title = {Levelling the Playing Field: A Comprehensive Comparison of Visual Place Recognition Approaches under Changing Conditions},
	booktitle = {IEEE ICRA Workshop on Dataset Generation and Benchmarking of SLAM Algorithms for Robotics and VR/AR},
	author = {Zaffar, Mubariz and Khaliq, Ahmad and Ehsan, Shoaib and Milford, Michael and McDonald-Maier, Klaus},
	month = apr,
	year = {2019},
    pages={1-8},
}

@INPROCEEDINGS{milford_seqslam_2012,
  address={Saint Paul, MN, USA},
  author={Milford, Michael J. and Wyeth, Gordon. F.},
  booktitle={2012 IEEE International Conference on Robotics and Automation}, 
  title={{SeqSLAM}: Visual route-based navigation for sunny summer days and stormy winter nights}, 
  year={2012},
  month=may,
  pages={1643-1649},
}

@INPROCEEDINGS{pepperell_all-environment_2014,
  address={Hong Kong, China},
  author={Pepperell, Edward and Corke, Peter I. and Milford, Michael J.},
  booktitle={2014 IEEE International Conference on Robotics and Automation (ICRA)}, 
  title={All-environment visual place recognition with {SMART}}, 
  month=may,
  year={2014},
  pages={1612-1618},
}

@incollection{pepperell_towards_2013,
    address={Sydney, Australia},
	title = {Towards persistent visual navigation using {SMART}},
	booktitle = {Proceedings of the 2013 Australasian Conference on Robotics and Automation},
	author = {Pepperell, Edward and Corke, Peter and Milford, Michael},
    month=dec,
	year = {2013},
    pages={1-9},
}

@inproceedings{garg_where_2021,
  address={Montreal, Canada},
  title     = {Where Is Your Place, Visual Place Recognition?},
  author    = {Garg, Sourav and Fischer, Tobias and Milford, Michael},
  booktitle = {Proceedings of the Thirtieth International Joint Conference on Artificial Intelligence (IJCAI)},
  pages     = {4416--4425},
  year      = {2021},
  month     = aug,
}

@ARTICLE{engel_direct_2018,
  author={Engel, Jakob and Koltun, Vladlen and Cremers, Daniel},
  journal={IEEE Transactions on Pattern Analysis and Machine Intelligence}, 
  title={Direct Sparse Odometry}, 
  month=mar,
  year={2018},
  volume={40},
  number={3},
  pages={611-625},
}

@INPROCEEDINGS{kim_scan_2018,
  address={Madrid, Spain},
  author={Kim, Giseop and Kim, Ayoung},
  booktitle={2018 IEEE/RSJ International Conference on Intelligent Robots and Systems (IROS)}, 
  title={Scan Context: Egocentric Spatial Descriptor for Place Recognition Within {3D} Point Cloud Map}, 
  month=oct,
  year={2018},
  pages={4802-4809},
}

@ARTICLE{matthies_error_1987,
  author={Matthies, L. and Shafer, S.},
  journal={IEEE Journal on Robotics and Automation}, 
  title={Error modeling in stereo navigation}, 
  month=jun,
  year={1987},
  volume={3},
  number={3},
  pages={239-248},
}

@article{maddern_1_2017,
	title = {1 year, 1000 km: The {Oxford} {RobotCar} dataset},
	volume = {36},
	number = {1},
	journal = {The International Journal of Robotics Research},
	author = {Maddern, Will and Pascoe, Geoffrey and Linegar, Chris and Newman, Paul},
	month = jan,
	year = {2017},
	pages = {3--15},
}

@inproceedings{sunderhauf_are_nodate,
  address={Karlsruhe, Germany},
  title={Are We There Yet? Challenging {SeqSLAM} on a 3000 km Journey Across All Four Seasons},
  author={S{\"u}nderhauf, Niko and Neubert, Peer and Protzel, Peter},
  booktitle={IEEE ICRA Workshop on Long-Term Autonomy},
  year={2013},
  month=may,
  pages={1-3},
}

\end{document}